\documentclass[pdflatex,sn-mathphys-num]{sn-jnl}

\usepackage{graphicx}%
\usepackage{multirow}%
\usepackage{amsmath,amssymb,amsfonts}%
\usepackage{amsthm}%
\usepackage{mathrsfs}%
\usepackage[title]{appendix}%
\usepackage{xcolor}%
\usepackage{textcomp}%
\usepackage{manyfoot}%
\usepackage{booktabs}%
\usepackage{algorithm}%
\usepackage{algorithmicx}%
\usepackage{algpseudocode}%
\usepackage{listings}%

\usepackage{comment}

\begin{document}

\title{Disjointness Violations in Wikidata}

\author*[1]{\fnm{Ege Atacan} \sur{Doğan}}\email{egeatacandogan@gmail.com}
\author*[2]{\fnm{Peter F.} \sur{Patel-Schneider}}\email{pfpschneider@gmail.com}
\affil*[1]{\orgdiv{Faculty of Engineering and Natural Sciences}, \orgname{Sabancı University}, \city{Istanbul}, \country{Turkey}}
\affil*[2]{\city{New Jersey}, \country{U. S. A.}}

\abstract{Disjointness checks are among the most important constraint checks in a knowledge base and can be used to help detect and correct incorrect statements and internal contradictions. Wikidata is a very large, community-managed knowledge base. Because of both its size and construction, Wikidata contains many incorrect statements and internal contradictions. We analyze the current modeling of disjointness on Wikidata, identify patterns that cause these disjointness violations and categorize them. We use SPARQL queries to identify each ``culprit'' causing a disjointness violation and lay out formulas to identify and fix conflicting information. We finally discuss how disjointness information could be better modeled and expanded in Wikidata in the future.

This is an extended version of a paper presented at the Sixth International Knowledge Graph and Semantic Web conference, December 2024.
}

\keywords{Wikidata, Knowledge Graph, Disjointness, Constraints}

\maketitle

\section{Introduction}\label{sec1}

Public knowledge graphs can be extended or improved in several ways. While adding new data to a knowledge graph is often seen as the primary method of contribution, as evidenced by the many web pages on how to add new data to Wikidata, managing internal consistency and developing tools for better user experience are also crucial. One method for increasing ease for users and ensuring internal consistency is through constraints applied either at edit time, with feedback to users, or run later via queries or external programs, producing constraint violation reports that users can employ to find and fix problems.

Wikidata \cite{wikidata} is the largest freely-editable knowledge graph, containing over 113 million objects (called {\em items} in Wikidata) as of the end of August 2024.  Wikidata encourages experts and non-experts alike to contribute.  In order to help maintain good internal consistency, Wikidata has constraints of various sorts, and there are dedicated communities that query errors and fix them.  In almost all areas, information in Wikidata is acknowledged to be incomplete with respect to the parts of the real world that it is modelling and this incompleteness contributed to the design of Wikidata.

An essential part of Wikidata is its simple but large, deep, wide, multi-domain, and foundational ontology of classes, where a class\footnote{www.wikidata.org/wiki/Wikidata:WikiProject\_Ontology/Classes} is a grouping of objects with common characteristics---the class's instances.   Wikidata classes are organized in a generalization (or subsumption) taxonomy with classes being subclasses of others.  Nearly all Wikidata classes are designed to be non-empty, i.e., even though there might not be any instances of the class in the current Wikidata knowledge graph there are objects in the real world (that might or not might not be present in Wikidata) that should belong to the class.

Any large repository of information has issues, and Wikidata, partly because it has been edited by many people, is no exccption.  There have been multiple investigations of issues in Wikidata, some related to Wikidata as a whole \cite{wikidata-quality-study} and some related to issues particularly in the Wikidata ontology \cite{wikidata-barriers, wikidata-ontology-issues-prioritization, wikidata-ontology-issues-solutions}.

One aspect of an ontology that can help to both find and reduce problems in a knowledge graph is disjointness between classes.  If two classes are known to be disjoint, given the presence of good editing tools, users will be warned when trying to put an object into disjoint classes or create a subclass of two disjoint classes, reducing the number of errors that end up in the knowledge graph.  As well, reports listing occurrences of the above two situations can be used to fix errors related to disjointness.

This paper examines the role disjointness plays and can play in helping maintain consistency in Wikidata.  It describes the current disjointness situation in Wikidata. It examines current issues with disjointness in Wikidata---finding violations and their sources.  It describes some reasons why incorrect information causing disjointness violations have ended up in Wikidata. It finally makes suggestions on how to better improve disjointness in Wikidata.
The work described in this paper was performed as part of a larger effort to find and fix issues in the Wikidata ontology. 

\section{Disjointness}\label{sec2}

In representation theory, two or more classes are pairwise disjoint when any two of the classes cannot have any instances in common. Two or more classes are mutually disjoint when there is no common instance for all the classes.  A (pairwise, mutual) disjointness statement is the assertion that two or more classes are (pairwise, mutually) disjoint. Disjointness statements are often used as constraints within knowledge graphs to help ensure that the knowledge graph faithfully represents the real world, preventing the creation of classes or items that are not part of the real world.  Disjointness statements also provide negative information, i.e., that an item is not an instance of some class because it is an instance of a disjoint class, and knowledge graphs generally are poor at representing negative information.

Wikidata has an unusual method for expressing disjointness.
Instead of having disjointness built into its language and
then having processes that enforce disjointness statements,
disjointness is expressed as regular statements (in the form of disjoint unions saying that
a class is the pairwise disjoint union of several other classes)\footnote{Disjoint union statements in Wikidata are a complete, partitioning, disjoint categorization relation in sense of
Almeida {\em et al} \cite{multi-level}, although there is no requirement in Wikidata that all the classes be in the same conceptual level.}
with no inferential support.
External processes can then use these statements to either generate reports
on disjointness violations in general or determine whether a particular item in Wikidata violated a disjointness statement.
It is up to Wikidata users who see these reports to determine what to do to fix the issue, if anything.
Wikidata than can and does end up with many disjointness violations.

In large ontologies like Wikidata's, any two randomly selected classes are more likely than not to be disjoint in the real world. This is particularly true for ontologies, like the Wikidata ontology, that cover many domains.  
So it might seem that many disjointness statements are required to express this large amount of disjointness.
However, because the ontology in Wikidata is large, deep, and wide, some disjointness statements can induce many disjointnesses. This happens in part because a pairwise disjointness statement with $n$ classes induces $n(n-1)/2$ disjointnesses. Another reason is that disjointness statements near the top of the Wikidata ontology have a much greater effect because if two classes are disjoint then each subclass of the first class is disjoint from each subclass of the second class.

As Wikidata includes a foundational ontology it has a universal class---``entity'' (Q35120)\footnote{Wikidata uses internal identifiers, here Q35120, and labels in multiple languages, here ``entity'' in English.  The labels are not guaranteed to be unique, even in a single language.  We will largely ignore identifiers outside of queries and only use the English label in double quotes.}---as the top class in its ontology.  (Information about ``entity`` can be seen at the Wikidata page \url{www.wikidata.org/wiki/Q35120}.)
Every item is an instance of ``entity'', and every class is a subclass of ``entity''.   
Disjoint union of statements
at the level of ``entity'' can have a very large effect, for example separating all the 1,340,122 subclasses of ``abstract entity'' from all the 2,775,101 subclasses of ``concrete object''.  (It turns out that this disjointness in Wikidata has about 47 thousand violations that have to be remedied before confidently making a statement of this form.)  Other disjoint union statements at the level of ``entity'', such as between ``observable entity'' and ``unobservable entity'', similarly have large effects.  On the other hand, most disjoint union statements in Wikidata are on very specific classes like ``first Monday of the month''.

\section{Methodology}\label{sec3}

Gathering disjointness information in Wikidata starts by writing SPARQL queries\footnote{We use SPARQL to access Wikidata throughout because of its flexibility and power, even though some simple accesses could have used other methods to access Wikidata.} against Wikidata encoded in RDF to find the pairwise disjoint classes in Wikidata disjoint union statements.  The queries used in this paper
were run during July and August of 2024 using the Wikidata Query service based on the Blazegraph SPARQL query engine \cite{BlazeGraph} at \url{query.wikidata.org}
when possible and the QLever Wikidata query service \cite{QLever} at \url{qlever.cs.uni-freiburg.de/wikidata}
when not.  (QLever is dramatically faster than Blazegraph on many queries but uses a weekly off-line dump.)  The query that returns these pairs is\footnote{The queries given here only report Wikidata IDs.  The actual queries report English labels as well and in QLever need prefix declarations---these parts of the queries are formulaic and not provided here.}

{\small
\begin{quote}
\begin{verbatim}
SELECT DISTINCT ?class ?e1 ?e2 WHERE {
  ?class p:P2738 ?l .
  MINUS { ?l wikibase:rank wikibase:DeprecatedRank . }
  ?l pq:P11260 ?e1 .
  ?l pq:P11260 ?e2 .
  FILTER ( ( str(?e1) < str(?e2) ) )
} ORDER BY ?class   
\end{verbatim}
\end{quote}
}

\noindent
The query returns triples of a class ID that has a non-deprecated ``disjoint union of'' (P2738) statement on it and pairs of class IDs that are stated to be pairwise disjoint in the disjoint union using ``list item'' (P11260) qualifiers.
There were 758 disjoint union statements on 631 classes resulting in 7,027 pairwise disjoint statements.

Although the above query is not too complex, it uses qualifiers, an advanced concept in Wikidata, and ignores some of the information needed to create a disjoint union.  The required use of qualifiers when specifying disjointness makes the query harder to write than it could be and definitely harder for users to create.  Worse, the requirement to have a class that is the disjoint union results in the creation of some classes in Wikidata just to be the disjoint union. Allowing direct disjointness statements in Wikidata makes it easier for users to specify disjointness.

Once the disjoint pairs have been retrieved, Wikidata can be queried to find violations of the constraints they place on Wikidata. Further querying can be performed to determine where a change needs to be made to fix the violation, producing {\em culprits} for disjointness violations. Select violations and culprits can be examined by hand to come up with probable explanations of why the incorrect information leading to the violation is in Wikidata.

There are also other classes in Wikidata that are intended to be disjoint with each other, such as the occupation classes with each other and with ``human''. Some of these disjointnesses are true in the real world but not reflected in any information in Wikidata.  The others can only be determined by consulting natural language text.  They are beyond the scope of this paper.

\section{Disjointness Violations in Wikidata}\label{sec4}

Once the disjoint pairs have been produced, a query can be evaluated for each pair finding issues related to the disjointness.  As mentioned above, a single disjointness between two classes can affect many other classes and many instances as well.  It is thus useful to not report every single issue ({\em violation}) resulting from a disjointness but instead only report the most important ones (the {\em culprits}).  Once the culprits have been identified the next step is to determine what mistake has been made and what change has to be performed to fix the problem.   For many culprits this is an inherently manual step that is outside the scope of this paper, but certainly part of the larger effort that this work is part of. 

\subsection{Scope}

The most important disjointness violations, and the ones examined here, are classes that are subclasses of both elements of a disjointness pair.  (Note that if the class is an empty class---in the real world, not just in Wikidata---then this is only a technical violation and not a true problem.)  There are also disjointness violations where an item is an instance of both elements of a disjointness pair.   We concentrate our examination on violations related to subclasses but do count all violations to get an idea of the size of the problem.

We define the subclass violations for a pair of disjoint classes, class1 and class2, as
\begin{align*}
\{ \text{class} \mid (\text{class} \subseteq \text{class1}) \wedge (\text{class} \subseteq \text{class2}) \} 
\end{align*}
\noindent
We define the instance violations for a pair of disjoint classes, class1 and class2, as 
\begin{align*}
 \{ \text{item} \mid (\text{item} \in \text{class1}) \wedge (\text{item} \in \text{class2}) \}
\end{align*}

\noindent
(In this paper several mathematical symbols are used to make formulae more readable.  Aside from the usual logical symbols, $\in$ is used for instance of, $\perp$ is used for disjoint, $\subseteq$ is used for subclass, and $\subset$ is used for direct subclass.)
\begin{figure}
    \centering
    \includegraphics[width=0.5\linewidth]{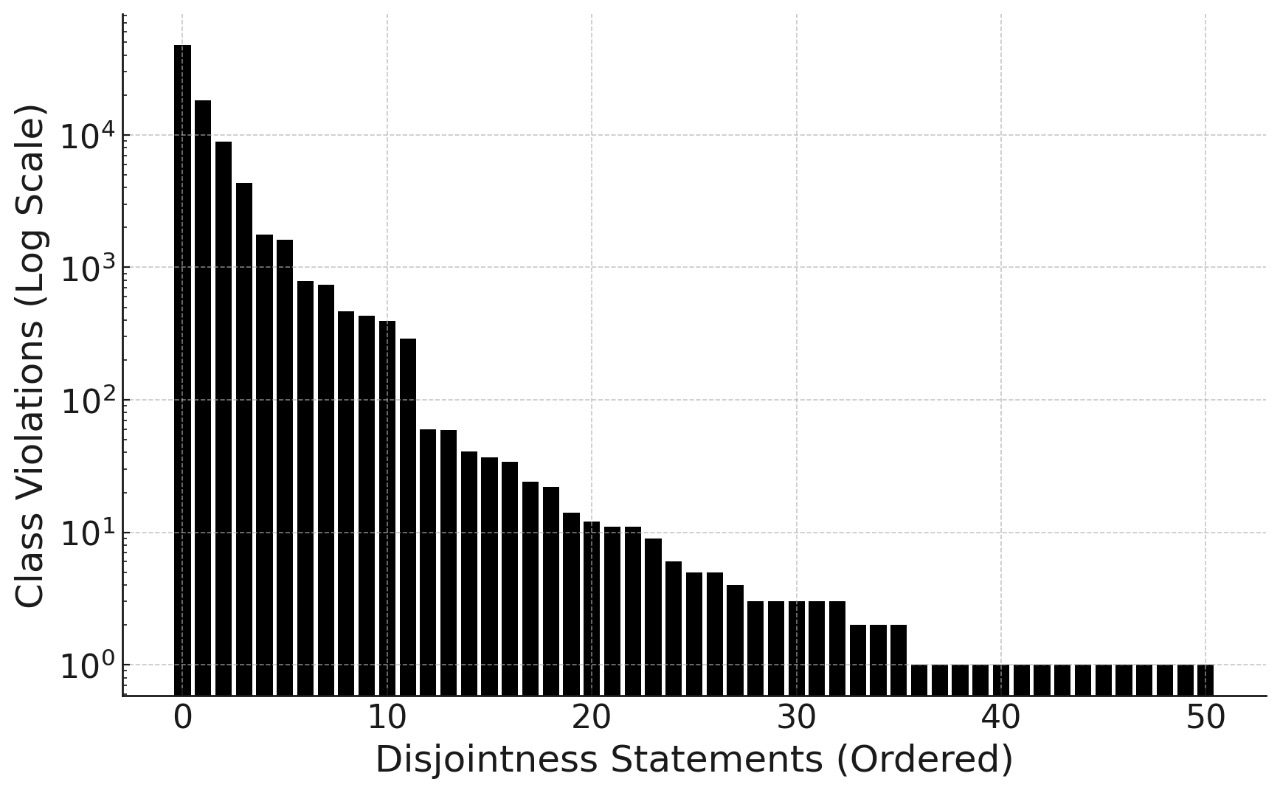}
    \caption{Class Violations per Disjointness Statement (log scale)}
    \label{fig:class}
\end{figure}

\begin{figure}
    \centering
    \includegraphics[width=0.5\linewidth]{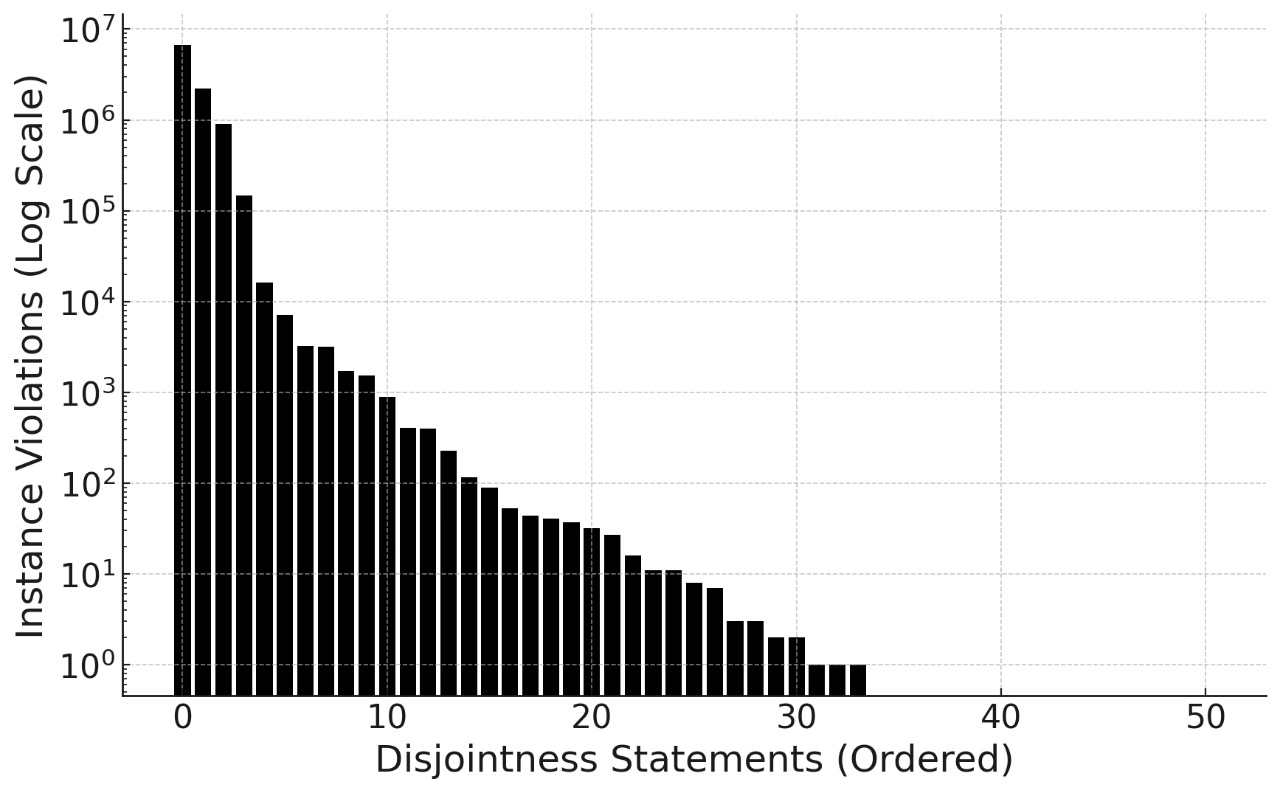}
    \caption{Instance Violations per Disjointness Statement (log scale)}
    \label{fig:instance}
\end{figure}

We used a Python program that first extracts all the disjointness pairs as above and then constructs queries that counts the number of subclass and instance violations for each pair.  
We ran this program in late August 2024, finding only 51 disjointness pairs with subclass violations, 
ranging from $47\,623$ subclass and $2\,203\,817$ instance violations for the disjointness between ``abstract entity'' and ``concrete object''
to $1$ subclass violation for several disjointnesses
and no instance violations for some disjointnesses.
Figures \ref{fig:class}~and~\ref{fig:instance} show the number of class and instance violations per disjointness statement, in reverse order.

Adding the number of violations for each disjointness pair results in
$86\,042$ subclass violations,  
$9\,951\,333$ instance violations,
and
$10\,037\,375$ total violations.

\subsection{Culprits}\label{subsec6}

\begin{figure}
    \centering
    \includegraphics[width=0.5\linewidth]{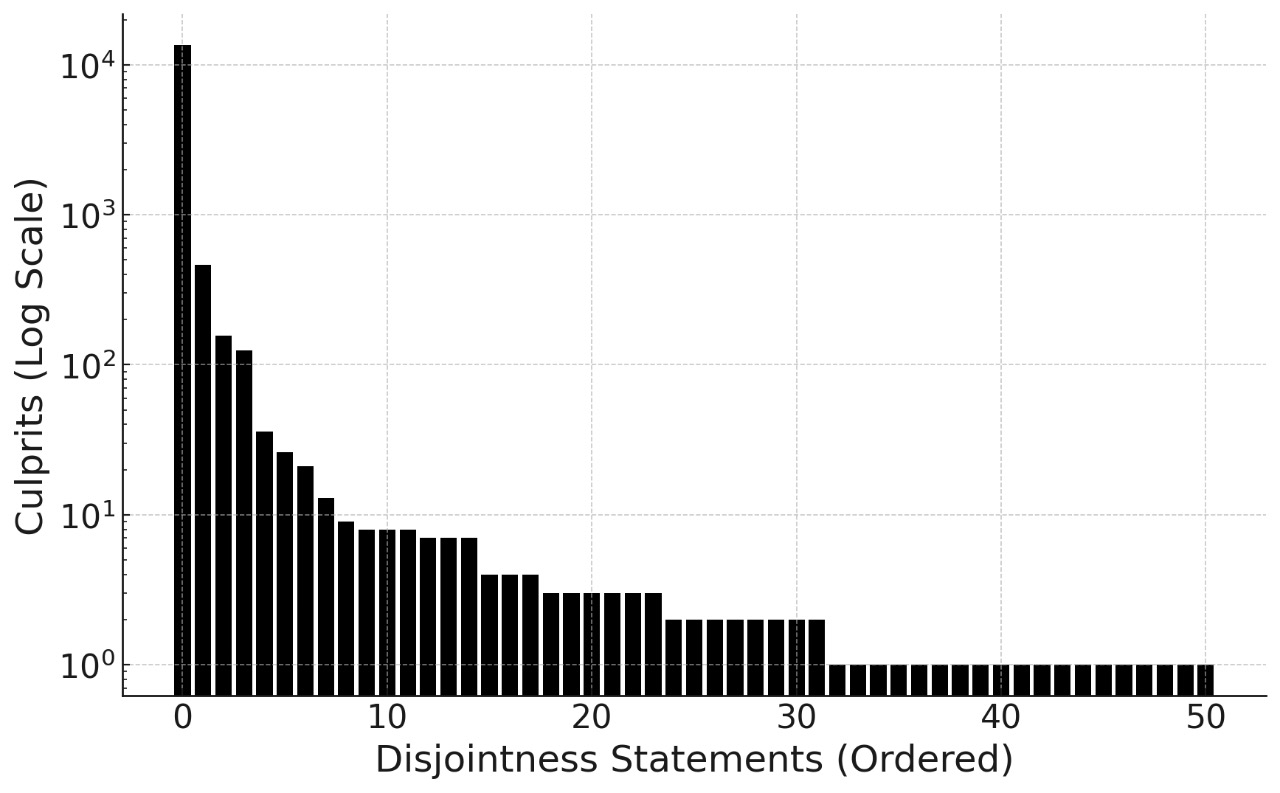}
    \caption{Culprits per Disjointness Statement (log scale)}
    \label{fig:culprit}
\end{figure}

The above numbers are large, with the number of instance violations much too large to be examined by hand, even with a large community involved.
It is thus useful to look for the violations that in some sense are the root causes of other subclass violations.  

If a class $A$ forms a disjointness violation because it is a subclass of both classes in a disjointness pair,
then all its subclasses will also be disjointness violations because they are
also subclasses of both of the disjoint classes.
Fixing the violation for $A$ will likely fix the disjointness violations for its subclasses.
It is thus useful to first look at the most-general classes
that form disjointness violations for each disjointness pair.
We call these most-general classes {\em culprits}.

For example,
``bow'' is a culprit because it is a subclass of both elements of the disjointness pair
``gun'' and ''draft weapon'' and none of its superclasses are subclass of both of the disjoint classes.
The violations for the 26 subclasses of ``bow'' will very likely be fixed if the violation on ``bow'' is fixed.

We formally define the culprits of a disjoint pair of classes, class1 and class2, as
\vspace*{-0.25ex} 
\begin{align*}
\{ \text{class} \mid & ( \text{class} \subseteq \text{class1} ) \wedge ( \text{class} \subseteq \text{class2} ) \; \wedge \\
& \nexists \text{parent} : (\text{class} \subset \text{parent}) \wedge (\text{parent} \subseteq \text{class1}) \wedge (\text{parent} \subseteq \text{class2}) \}
\end{align*}

Our Python program also counted culprits for each disjointness pair.
As of late August 2024, $14\,480$ total culprits existed on Wikidata.
Again the number of culprits per disjointness pair varies widely, as shown in Figure~\ref{fig:culprit},
ranging from $13\,520$ for the disjointness between ``abstract entity'' and ``concrete object''
to $1$ for several disjointnesses including the one between ``phonograph record'' and ``compact disc''.

A quick perusal of the culprits showed many ($10\,753$ by checking subclass) related to ``gene'' and being a violation of the disjointness between ``abstract entity'' and ``concrete object''.  
On average, each culprit causes almost 6 subclass violations and $687$ violations in total.
Averages do not give much information about any given culprit here, as the distribution is again very irregular.

The number of culprits shows that, while disjointness is a very frequently violated constraint on Wikidata, the required fix is likely much smaller in scope than the problem itself.
The set of culprits is small enough that it can be exhaustively examined by hand by a small community.

Note that some items can be culprits more than once. For example, ``Turkmen tribes''is a culprit in regards to two separate disjointnesses (``abstract entity'', ``concrete object''  and ``person'', ``organization'') and thus is counted twice. Counting these items more than once is beneficial, as they may be pointing to more than one mistake.

\subsection{Mistakes and Fixes}\label{subsec7}

Identifying culprits is only an intermediate step to the goal of fixing disjointness issues.
The final step is to determine what mistake was made and how to fix it.
Determining mistakes cannot be done with a simple formula and then retrieved using a simple query like determining culprits.
Instead, identifying and fixing the mistake that lead to a culprit (or a disjointness violation in general) requires manually examining the information in Wikidata, often using one of the following methods:

\begin{itemize}
    \item Check if many similar items are culprits, as is the case for the culprits under ``gene''. The mistake is likely then higher up instead of the culprit. Fix it when found.
    \item Check if the disjointness relationship that is being violated makes sense. If it doesn’t, check its violations to get an idea about what might be wrong, change or remove it.
    \item Try to decide whether the class might be an empty class, if so, make it a subclass of ``the empty class'', and thus give a reason why the violation is not a problem.
    \item If not any of the cases above, try to understand what was intended. The violation could be caused by one of the sources of editing errors mentioned later in the paper, which then points to a way of fixing the violation.
    \item Sometimes the mistake or fix cannot be determined easily.  In these cases it is often best to ask people from the part of the Wikidata community active in the domain of the culprit class.
\end{itemize}

%
In any systematic investigation of culprits, it is useful to record how many mistakes were identified and how many culprit violations their fix eliminated by checking after each fix which culprits still are disjointness violations.
This check can only be approximate as other changes to Wikidata might also fix disjointness violations.\footnote{We have not observed any other changes to Wikidata that have fixed any significant number of disjointness violations.}
If only some violations are eliminated this information can be used to show the effectiveness of the work done so far and maybe help with the rest of the process.
If all the culprit violations are eliminated this information can be used to show how the culprits are clustered, and maybe point to better ways of identifying related culprits that may then lead to better ways of identifying mistakes and fixes for future rounds of disjointness violation elimination.

\subsection{Example}

We have only started the work on identifying mistakes so we have only a little data on effectiveness. Our identification of mistakes started with culprits that we strongly suspected would lead to mistakes and fixes that eliminate many culprit disjointness violations.  We provide here an examination of the most prominent mistake, one that affects the majority of culprits.

As mentioned above, many culprits were related to ``gene''.   
The underlying issue was that ``gene'' was used to represent both the sequence of bases, which is an abstract piece of data, and the physical molecules that exist within organisms. 
These two concepts should be represented by separate items. 
There was no disjointness violation on ``gene'' itself because it itself was not a subclass of ``concrete object'' so it was not considered a culprit.
Instead many of the subclasses of ``gene'' are also subclasses of ``concrete object'',
including $10\,656$ that were identified as culprits using the following SPARQL query:

\begin{verbatim}
SELECT (COUNT(DISTINCT ?class) AS ?count) WHERE {
  { SELECT  ?class WHERE {
      ?class wdt:P279+ wd:Q7187 .
      ?class wdt:P279+ wd:Q4406616 .
  } }
  MINUS {
    ?class wdt:P279 ?parent .
    ?parent wdt:P279+ wd:Q7187 .
    ?parent wdt:P279+ wd:Q4406616 .
} }  
\end{verbatim}

Looking at the culprits and the Wikidata ontology around them and ``gene'' (Q7187) indicates that the mistake is that ``gene'' is a subclass of ''abstract object'' (Q4406616), as genes are physical.  The fix is thus removing ``subclass of'' (P279) links to make ``gene'' no longer being a subclass of ``abstract entity''.   We haven't made this change so as to not interfere with other parts of our investigations but it will fix the disjointness violations for all of the culprits identified above.

This mistake was by far the most prevalent among the culprits, very easily noticeable in the culprits table provided with the extended version of the paper.  There are 3727 other culprits, so we have shown that the number of total mistakes is smaller than or equal to 3728.

\section{Kinds of Disjointness Violations}\label{sec5}

In our examination of the culprits we have identified several kinds of violations based on where the mistake occurs in relation to the culprit and whether there is really a mistake at all.

\paragraph{Local mistakes}
In many cases, there is simply some sort of mistake on the culprit that needs to be fixed to eliminate the violation.  
For example, ``linguistic rights activist'' is a culprit because it is subclass of both ``abstract entity'' via ``social movement'' and ``concrete object`` via ``political activist''.
The first subclass relationship is incorrect and removing the link eliminates the disjointness violation.

\paragraph{Mistakes in superclasses}
In other cases the information on the culprit is locally correct but there is a superclass of the culprit that has a mistake that needs to be fixed.  
For example, as shown above ``gene'' is a superclass of many culprits but is not a culprit itself.  Fixing the incorrect superclass of ``gene'' dramatically reduces the number of culprits.

\paragraph{Incorrect disjointnesses}
Our work here starts with the assumption that the disjointness pairs are correctly disjoint but it may be that the disjointness itself is incorrect and either needs to be removed or modified.  For example, ``vehicle'' is stated to be the disjoint union of ``land vehicle'', ``watercraft'', ``aircraft'', and ``spacecraft''.  This is not a correct partition because there are vehicles that can belong to several of these categories, such as ``water-based aircraft''.  Resolving this mistake can be done by either removing the disjoint union or adding an extra possibility for mixed-area vehicles.  The latter solution would require adjusting many other classes, so the former is likely the preferred solution.  


\paragraph{Empty classes} 
An empty class is a class that cannot have any instances. (This is not the same as a class which only has fictional instances, such as ``unicorn''.) As part of our effort here, we have done some work on representing empty classes on Wikidata.\footnote{We have added the ``empty class'' class (Q128139417), and the ``the empty class'' item (Q126726396). The first is the class containing all empty classes (and therefore is not an empty class itself). The second is a class that is empty, it is equivalent to all other empty classes, and also subclass of every class. Empty classes were not on Wikidata before we did this.  It may be counterintuitive for a knowledge graph to have empty classes, and the vast bulk of classes in Wikidata are not empty.  However, within foundational ontologies, all bits of information can be represented. (This is in line with the goal of Wikimedia projects to contain the ``sum of all human knowledge''.) The fact that a certain class does not have any instances is important to know, many disciplines tackle the question of ``Does x exist?'', and the results are worthy of being in a knowledge graph.} An example of an empty class is ``abnormal number'', as it is mathematically impossible for anything to be an instance of this class.

If classes can be empty  
a disjointness violation on a class does not necessarily mean that there is an error,
just that the class, and all its subclasses, are necessarily empty classes. As in Wikidata it is rarely the case that a class is empty, and the more likely occurrence is that an error has been made.  Our view is that the best solution is to require that all empty classes are stated to be instances of ``empty class''. 

So the fix to the kind of mistake that involves an empty class is to have the class be an instance of ``empty class''.  (Once this is being done queries and tools that identify errors relating to disjointness should take into account the possibility of a class being an instance of ``empty class''.)
Note that some classes are not empty by definition, but just currently don't have any instances in Wikidata \cite{learning-disjointness}. These classes do not fit into the empty class category.

\section{Sources of Issues with Disjointness}\label{sec6}

The above examinations point out and quantify some of the characteristics of disjointness violations as they occur in Wikidata.  But they do not directly speak to why this incorrect information is present in Wikidata, i.e., what is the reason that Wikidata users added information to Wikidata that created disjointness violations.

Recently the Wikidata community and Wikimedia Deutschland conducted a survey and several discussions on issues in the Wikidata ontology \cite{wikidata-ontology-survey-prioritization, wikidata-ontology-issues}
and some potential solutions were proposed
\cite{wikidata-ontology-issues-solutions}.
The issues here closely mirror some of the major issues in this previous work.
Disjointness violations and the work here provide a way of uncovering many examples of ontology issues, the start of any largbe-scale attempt to improve the Wikidata ontology.

\paragraph{Exceptions} 
One observed cause of disjointness violations is exceptions, where a class $x$ is a subclass of $y$ which provides a characteristic for instances of $x$ but a subclass of $x$ has a different characteristic that is provided by a superclass stated to be disjoint from $y$. For example, ``lake'' is a subclass of ``natural object'' but its subclass ``artificial lake'' is itself a subclass of ``artificial object'' which is disjoint from ``natural object''.

A possible solution is to have a subclass of ``lake'' for natural lakes and move the ``natural object'' superclass away from ``lake'' to natural lakes.  Another solution is to remove the subclass relationship between human-made lake and lake, as it can be argued that human-made lakes somehow are not ``real lakes'', but something else, and keeping the same superclasses for ``lake''.

\paragraph{Ambiguous labels}\label{subsec10}
Many users of Wikidata determine the meaning of a class almost solely from its label (and sometimes also from its description).  But natural language, particularly short phrases, is often ambiguous resulting in superclasses for several of the meanings of the label (or description) and this can lead to the class having disjoint superclasses.  For example, ``food waste'' can refer to either food that was wasted, and thus a subclass of ``concrete object'' via ``biodegradable waste'', or an act of wasting food, and thus a subclass of ``abstract entity'' via ``waste of resources''.   As ``concrete object'' and ``abstract entity'' are disjoint ``food waste''  cannot be a subclass of both.  

A possible solution is to split ``food waste'' into two classes.  Another solution is to determine which reading is correct based on other information about the class, such as its descriptions or instances, and remove the incorrect superclass.

\paragraph{Multiple senses of a word}\label{subsec11}

A related issue arises from words that have multiple meanings.  Similarly to the above case, the different meanings for a class whose label is a single word can give rise to disjoint superclasses for the class.   For example, ``foil'' is a subclass of both  ``concrete object'' (via ``ornament'') and ``abstract entity'' (via ``motif''). Foil is used both as the material that is used in art, and also as an artistic part of the work. Another example is ``disease'', that leads to a class order related confusion (see just below) as it both means strain of disease, and the disease that someone specifically has.

\paragraph{Class order confusion}\label{subsec12}

Wikidata allows classes to be instances of (higher order) classes.  This induces a hierarchy of classes.   Some of these classes are fixed order, where a first-order class has only non-classes as instances, a second-order class has only first-order classes as instances, etc. All fixed order classes are (pairwise) disjoint with each other. As with other parts of Wikidata, there are issues in this hierarchy, particularly where classes that should be instances of another class are instead a subclass of the other class, or vice versa \cite{wikidata-conceptual-disarray}.

For example, ``chemical element'' is stated to be a second-order class but is also a subclass of ``concrete object'', which is stated to be a first-order class.  As second-order classes are (correctly) stated to be disjoint from first-order classes, this produces a disjointness violation.  The solution here is to remove ``chemical element'' from the subclasses of ``concrete objects'' as chemical elements, e.g., ``Mercury'' is a class with concrete object instances, it is not a concrete object itself.

\paragraph{Mixture}\label{subsec13}

Wikidata has classes whose instances are mixtures of several components.  In some cases the class is a subclass of the component classes.  If the component classes are disjoint then a disjointness violation results.  For example ``BBP DANSAERT Saison x Lambic'' is a subclass of both ``saison'' and ``lambic'', two classes of beer that are subclasses of two disjoint beer characteristic classes (``high fermentation beer'' and ``spontaneous fermentation beer'').  

There are two problems here that are typical of mixtures.  First, a mixture class should have the component classes as parts, not superclasses.    Second the disjointness comes from ``beer'' being the disjoint union of several characteristic classes with no allowance for beers that mix several different characteristics. An extra member of the disjoint union is needed to account for such mixtures, adding that extra member satisfies the exhaustive partition criteria of a disjointness statement.

The similar problem shows up in situations that are not physical mixtures, but where a combination of characteristics is somehow possible.  The class ``game'' is the disjoint union of ``game of chance'', ``game of skill'', and ``game combining chance and skill'', explicitly allowing for combinations.

\paragraph{Confusing items}\label{subsec14}

For some classes their meaning is not ambiguous but how to categorize the class is unclear.  The class may have multiple aspects that fit under disjoint superclasses.  For example, ``biological sequence'' was described as several small chemical fragments (monomers) linked together into a polymer with biological utility.   One aspect of instances of the class is their physicality, leading via ``biomolecular structure'' to the superclass ``concrete object''; another is the arrangement of monomers, leading via ``sequence'' to the superclass ``abstract entity''.  Together these result in a disjointness violation.

The solution for this sort of class is to determine which aspect is more fitting and remove the other, possibly creating a new class for this other aspect and link the two classes together. This kind of splitting of classes can be contentious and generally needs to be discussed within the Wikidata community. A safe way to do this split is to cross-check from other databases and/or knowledge graphs.

The disjointness violation on ``biological sequence'' was causing around fifty thousand subclass violations so we used it as a test case of how to eliminate violations before we did the bulk of our analysis.
Because BioPortal's ontology \cite{bioportal-edam} considers the equivalent class a subclass of ``data'' we marked ``biological sequence'' as a subclass of ``data''. Since data is subclass of ``abstract entity'' we decided to keep ``sequence'' as a superclass and to remove ``biomolecular structure''. We also cited Bioportal's ontology as a reference for new link to ``data''.  Information in Wikidata is generally supposed to be supported by references so having solid references reduces potential pushback from the community.

\paragraph{Basic mistakes}\label{subsec15}

Some disjointness violations just appear to be the result of some misreading of either the label or description of a class.  For example, ``linguistic rights activist'' is subclass of ``social movement''. While a class for linguistic rights activism would indeed be a subclass of ``social movement'', the person who is an activist cannot be considered an instance of a social movement.  These sorts of basic mistakes are generally fixable by just removing the incorrect subclass statement.

\paragraph{Vandalism}\label{subsec16}

Vandalism is a problem with every freely-editable shared resource.  Wikidata has had problems with vandalism \cite{vandalism} and it is possible that a disjointness violation is the result of vandalism on an item.  However, such occurrences are infrequent on Wikidata, as other shared resources, such as Wikipedia, are vandalized more often.

\bigskip

Many of the above issues come from a lack of examination of the relevant parts of the Wikidata ontology.  For example, looking at the ancestors of a class in the ontology would both reveal the characteristics of a class (such as being natural) and which of several readings of a label or description is correct. 

\section{Suggestions for Improving Disjointness}\label{sec8}

Even though Wikidata has a significant number of disjointnesses because of the far reach of some of its disjointness statements, disjointness could play a larger positive part in Wikidata. Just adding more disjointness statements to Wikidata, especially at the middle levels of the ontology, without making any other changes would provide this, but changes to Wikidata tools and to Wikidata itself would help increase the use of disjointness in Wikidata.

One possibility is to just go through parts of the Wikidata ontology looking for cases where a class appears to be a disjoint union of other classes and adding a ``disjoint union of'' statement to the class and then determining whether the disjointness is correct based on examining the new disjointness violations.
It is unclear, however, how effective this process would be if performed by users with limited domain knowledge, particularly if the violations can only be detected using QLever with the current week-long delay.

\subparagraph{\bf Better use of tools}
One problem with disjointness in Wikidata is that disjointness violations are not shown to users when they are editing Wikidata unless they use a special plugin---\url{www.wikidata.org/w/index.php?title=User%3ATomT0m%2Fclassification.js&action=raw&ctype=text%2Fjavascript}.  
Making this useful tool a part of the core user interface would help in showing disjointness violations to users.

Another problem is that reporting on disjointness violations cannot be done in the official Wikidata query service because of its poor performance.   Moving to a faster query service with performance on par with QLever would allow for real-time examination of disjointness violations.\footnote{As of August 2024 QLever only works on offline RDF dumps of Wikidata.}

\subparagraph{\bf Better disjointness constructs}
Disjointness in Wikidata is only possible as part of a disjoint union construct. This can require the creation of artificial classes to be the union, such as ``award, award nominees, award recipients or award ceremony'' (Q26877490).  The disjoint union construct also requires the use of qualifiers and an artifical item to be the value of the statement. Further, the disjoint classes have to be specified inside the construct, separate from any other information about them.

Adding a construct that, for example, made all the instances or direct subclasses of a class pairwise disjoint would eliminate the need for a union class and make disjointness easier for users to state. As well, there are existing classes, such as some of the classes of classes in the biological domain whose instances are disjoint, eliminating the need to separately state the disjoint classes. This could be used, for example, to state that the different mammalian species are disjoint.

\subparagraph{\bf Crowdsourcing fixes}
To effectively reduce the number of disjointness violations, particularly instance violations that are not covered by our culprits, it will likely be useful to employ the Wikidata community as a whole.  A tool that showed disjointness culprits in context and pointed out changes that would eliminate the issue would reduce the effort required to reduce disjointness violations and would consequently allow Wikidata to accomodate more disjointness statements that are regularly checked.  If the tool had game-like rewards \cite{wikipedia-gamifying} that might help with its uptake in the community.

\section{Summary and Future Work}\label{sec9}

Disjointness is a significant part of Wikidata.   Many classes can be inferred to be disjoint based on the ``disjoint union of'' statements in Wikidata, particularly those at the highest levels of the Wikidata ontology.  Unfortunately, there are many disjointness violations, where a class is a subclass of two disjoint classes or an item is an instance of two disjoint classes.  Fortunately, a much smaller but still large number of culprits cause these violations.

Several empirically determined causes give rise to many of these violations.  These causes appear to be mostly related to not taking into account information that helps define items when editing Wikidata, indicating that requiring confirmation for changes that create disjointness violations might induce users to investigate further, such as by looking at the superclasses of a class, fixing their erroneous edits before they adds to the number of disjointness violations.

We are continuing to fix disjointness violations.  The current state of our work is described at \url{www.wikidata.org/wiki/User:Egezort/Fixing_all_(yes_all)_Disjointness_Violations}. We have tackled most of the disjointness statements that produced a small number of violations, leaving the larger ones for later, perhaps as a community effort.

The ultimate goal of this line of work is to have Wikidata have a large and useful set of disjointness statements with no disjointness violations.  We plan to investigate a large sample of the culprits we have identified, fix the issues that resulted their violations, and see just how many other disjointness violations are resolved.  We also plan to work on the suggestions above for improving disjointness in Wikidata in order to drive towards this goal.

\section*{Acknowledgments}

Ege Atacan Doğan was partly supported by the ERASMUS Student Mobility for Traineeships program, hosted at the Free University of Bozen-Bolzano and supervised by Enrico Franconi.

\bibliography{disjointness}


\begin{thebibliography}{15}
\ifx \bisbn   \undefined \def \bisbn  #1{ISBN #1}\fi
\ifx \binits  \undefined \def \binits#1{#1}\fi
\ifx \bauthor  \undefined \def \bauthor#1{#1}\fi
\ifx \batitle  \undefined \def \batitle#1{#1}\fi
\ifx \bjtitle  \undefined \def \bjtitle#1{#1}\fi
\ifx \bvolume  \undefined \def \bvolume#1{\textbf{#1}}\fi
\ifx \byear  \undefined \def \byear#1{#1}\fi
\ifx \bissue  \undefined \def \bissue#1{#1}\fi
\ifx \bfpage  \undefined \def \bfpage#1{#1}\fi
\ifx \blpage  \undefined \def \blpage #1{#1}\fi
\ifx \burl  \undefined \def \burl#1{\textsf{#1}}\fi
\ifx \doiurl  \undefined \def \doiurl#1{\url{https://doi.org/#1}}\fi
\ifx \betal  \undefined \def \betal{\textit{et al.}}\fi
\ifx \binstitute  \undefined \def \binstitute#1{#1}\fi
\ifx \binstitutionaled  \undefined \def \binstitutionaled#1{#1}\fi
\ifx \bctitle  \undefined \def \bctitle#1{#1}\fi
\ifx \beditor  \undefined \def \beditor#1{#1}\fi
\ifx \bpublisher  \undefined \def \bpublisher#1{#1}\fi
\ifx \bbtitle  \undefined \def \bbtitle#1{#1}\fi
\ifx \bedition  \undefined \def \bedition#1{#1}\fi
\ifx \bseriesno  \undefined \def \bseriesno#1{#1}\fi
\ifx \blocation  \undefined \def \blocation#1{#1}\fi
\ifx \bsertitle  \undefined \def \bsertitle#1{#1}\fi
\ifx \bsnm \undefined \def \bsnm#1{#1}\fi
\ifx \bsuffix \undefined \def \bsuffix#1{#1}\fi
\ifx \bparticle \undefined \def \bparticle#1{#1}\fi
\ifx \barticle \undefined \def \barticle#1{#1}\fi
\bibcommenthead
\ifx \bconfdate \undefined \def \bconfdate #1{#1}\fi
\ifx \botherref \undefined \def \botherref #1{#1}\fi
\ifx \url \undefined \def \url#1{\textsf{#1}}\fi
\ifx \bchapter \undefined \def \bchapter#1{#1}\fi
\ifx \bbook \undefined \def \bbook#1{#1}\fi
\ifx \bcomment \undefined \def \bcomment#1{#1}\fi
\ifx \oauthor \undefined \def \oauthor#1{#1}\fi
\ifx \citeauthoryear \undefined \def \citeauthoryear#1{#1}\fi
\ifx \endbibitem  \undefined \def \endbibitem {}\fi
\ifx \bconflocation  \undefined \def \bconflocation#1{#1}\fi
\ifx \arxivurl  \undefined \def \arxivurl#1{\textsf{#1}}\fi
\csname PreBibitemsHook\endcsname

\bibitem[\protect\citeauthoryear{Vrande{\v{c}}i{\'c} and
  Kr{\"o}tzsch}{2014}]{wikidata}
\begin{barticle}
\bauthor{\bsnm{Vrande{\v{c}}i{\'c}}, \binits{D.}},
\bauthor{\bsnm{Kr{\"o}tzsch}, \binits{M.}}:
\batitle{Wikidata: A free collaborative knowledgebase}.
\bjtitle{Communications of the ACM}
\bvolume{57}(\bissue{10}),
\bfpage{78}--\blpage{85}
(\byear{2014})
\end{barticle}
\endbibitem

\bibitem[\protect\citeauthoryear{Shenoy et~al.}{2022}]{wikidata-quality-study}
\begin{botherref}
\oauthor{\bsnm{Shenoy}, \binits{K.}},
\oauthor{\bsnm{Ilievski}, \binits{F.}},
\oauthor{\bsnm{Garijo}, \binits{D.}},
\oauthor{\bsnm{Schwabe}, \binits{D.}},
\oauthor{\bsnm{Szekely}, \binits{P.}}:
A study of the quality of {W}ikidata.
Journal of Web Semantics
\textbf{72}
(2022)
\end{botherref}
\endbibitem

\bibitem[\protect\citeauthoryear{Patel-Schneider}{2019}]{wikidata-barriers}
\begin{bchapter}
\bauthor{\bsnm{Patel-Schneider}, \binits{P.F.}}:
\bctitle{Barriers to using {W}ikidata as a knowledge base}.
In: \bbtitle{WikidataCon 2019}.
\bpublisher{\url{www.wikidata.org/wiki/Wikidata:WikidataCon_2019}},
\blocation{Berlin}
(\byear{2019}).
\bcomment{Accessed 2024-07-23.}
\end{bchapter}
\endbibitem

\bibitem[\protect\citeauthoryear{Abdulai and
  Lacroix}{2023}]{wikidata-ontology-issues-prioritization}
\begin{botherref}
\oauthor{\bsnm{Abdulai}, \binits{M.}},
\oauthor{\bsnm{Lacroix}, \binits{L.}}:
Wikidata: Ontology Issues Prioritization.
\url{www.wikidata.org/wiki/Wikidata:Ontology_issues_prioritization}.
Accessed 2024-07-17.
(2023)
\end{botherref}
\endbibitem

\bibitem[\protect\citeauthoryear{Pintscher}{2023}]{wikidata-ontology-issues-solutions}
\begin{botherref}
\oauthor{\bsnm{Pintscher}, \binits{L.}}:
Wikidata Survey on Ontology Issues, Potential Solutions.
\url{www.wikidata.org/wiki/Wikidata_talk:Ontology_issues_prioritization#Overview_of_potential_solutions}.
Accessed 2024-07-17.
(2023)
\end{botherref}
\endbibitem

\bibitem[\protect\citeauthoryear{Almeida et~al.}{2018}]{multi-level}
\begin{bchapter}
\bauthor{\bsnm{Almeida}, \binits{J.P.A.}},
\bauthor{\bsnm{Carvalho}, \binits{V.A.}},
\bauthor{\bsnm{Brasileiro}, \binits{F.}},
\bauthor{\bsnm{Fonseca}, \binits{C.M.}},
\bauthor{\bsnm{Guizzardi}, \binits{G.}}:
\bctitle{Multi-level conceptual modeling: Theory and applications}.
In: \bbtitle{Proceedings of the XI Seminar on Ontology Research in Brazil and
  II Doctoral and Masters Consortium on Ontologies}
\bpublisher{ceur-ws.org/Vol-2228}, \blocation{???}
(\byear{2018})
\end{bchapter}
\endbibitem

\bibitem[\protect\citeauthoryear{}{2013}]{BlazeGraph}
\begin{botherref}
Welcome to Blazegraph.
\url{blazegraph.com}.
Accessed 2024-07-23.
(2013)
\end{botherref}
\endbibitem

\bibitem[\protect\citeauthoryear{Bast and Buchhold}{2017}]{QLever}
\begin{bchapter}
\bauthor{\bsnm{Bast}, \binits{H.}},
\bauthor{\bsnm{Buchhold}, \binits{B.}}:
\bctitle{{QLever}: A query engine for efficient {SPARQL}+text search}.
In: \bbtitle{CIKM '17: ACM Conference on Information and Knowledge Management},
\bconflocation{Singapore}
(\byear{2017})
\end{bchapter}
\endbibitem

\bibitem[\protect\citeauthoryear{V{\"o}lker
  et~al.}{2007}]{learning-disjointness}
\begin{bchapter}
\bauthor{\bsnm{V{\"o}lker}, \binits{J.}},
\bauthor{\bsnm{Vrande{\v{c}}i{\'{c}}}, \binits{D.}},
\bauthor{\bsnm{Sure}, \binits{Y.}},
\bauthor{\bsnm{Hotho}, \binits{A.}}:
\bctitle{Learning disjointness}.
In: \beditor{\bsnm{Franconi}, \binits{E.}},
\beditor{\bsnm{Kifer}, \binits{M.}},
\beditor{\bsnm{May}, \binits{W.}} (eds.)
\bbtitle{The Semantic Web: Research and Applications},
pp. \bfpage{175}--\blpage{189}.
\bpublisher{Springer},
\blocation{Berlin, Heidelberg}
(\byear{2007})
\end{bchapter}
\endbibitem

\bibitem[\protect\citeauthoryear{Abdulai}{2023}]{wikidata-ontology-survey-prioritization}
\begin{botherref}
\oauthor{\bsnm{Abdulai}, \binits{M.}}:
Wikidata: Ontology Issues Prioritization.
\url{www.wikidata.org/wiki/Wikidata:Ontology_issues_prioritization}.
Accessed 2024-07-23.
(2023)
\end{botherref}
\endbibitem

\bibitem[\protect\citeauthoryear{Pintscher}{2023}]{wikidata-ontology-issues}
\begin{bchapter}
\bauthor{\bsnm{Pintscher}, \binits{L.}}:
\bctitle{Ontology issues in {W}ikidata: {E}verything in neat and tidy boxes?
  {N}ot quite!}
In: \bbtitle{WikidataCon 2023}
\bpublisher{\url{commons.wikimedia.org/wiki/File:WikidataCon_2023_Ontology_issues_in_Wikidata_-_Everything_in_neat_and_tidy_boxes_Not_quite!.pdf}},
  \blocation{???}
(\byear{2023}).
\bcomment{Accessed 2024-07-23.}
\end{bchapter}
\endbibitem

\bibitem[\protect\citeauthoryear{Dadalto
  et~al.}{2021}]{wikidata-conceptual-disarray}
\begin{bchapter}
\bauthor{\bsnm{Dadalto}, \binits{A.A.}},
\bauthor{\bsnm{Almeida}, \binits{J.P.A.}},
\bauthor{\bsnm{Fonseca}, \binits{C.M.}},
\bauthor{},
\bauthor{\bsnm{Guizzardi}, \binits{G.}}:
\bctitle{Type or individual? {E}vidence of large-scale conceptual disarray in
  {W}ikidata}.
In: \bbtitle{40th International Conference on Conceptual Modeling (ER 2021)},
pp. \bfpage{367}--\blpage{377}
(\byear{2021}).
\doiurl{10.1007/978-3-030-89022-3_29}
\end{bchapter}
\endbibitem

\bibitem[\protect\citeauthoryear{{BioPortal}}{2024}]{bioportal-edam}
\begin{botherref}
\oauthor{\bsnm{{BioPortal}}}:
{EDAM Ontology}.
\url{https://bioportal.bioontology.org/ontologies/EDAM?p=classes&conceptid=data_2044}.
Accessed: 2024-08-28
(2024)
\end{botherref}
\endbibitem

\bibitem[\protect\citeauthoryear{Heindorf et~al.}{2016}]{vandalism}
\begin{bchapter}
\bauthor{\bsnm{Heindorf}, \binits{S.}},
\bauthor{\bsnm{Potthast}, \binits{M.}},
\bauthor{\bsnm{Stein}, \binits{B.}},
\bauthor{\bsnm{Engels}, \binits{G.}}:
\bctitle{Vandalism detection in {W}ikidata}.
In: \bbtitle{CIKM '16: Proceedings of the 25th ACM International on Conference
  on Information and Knowledge Management},
pp. \bfpage{327}--\blpage{333}
(\byear{2016})
\end{bchapter}
\endbibitem

\bibitem[\protect\citeauthoryear{Oceja and Sierra}{2018}]{wikipedia-gamifying}
\begin{bchapter}
\bauthor{\bsnm{Oceja}, \binits{J.}},
\bauthor{\bsnm{Sierra}, \binits{A.O.}}:
\bctitle{Gamifiying {W}ikipedia?}
In: \bbtitle{12th European Conference on Games Based Learning},
\bconflocation{Sophia Antipolis, France}
(\byear{2018})
\end{bchapter}
\endbibitem

\end{thebibliography}

\appendix

\section{Programs and Results}

The Python program AllCulprits.py in the ancillary files
first queries Wikidata to find the disjoint pairs in the disjoint union of statements
and then runs a query for each pair to find culprits for the pair.

The main results are in AllCulprits.csv, showing the two disjoint classes of the pair, the class that the disjoint union is on, and a culprit class for the pair.   Each class is shown as its identifier in Wikidata and its English label, if any.
The violations between ``concrete object'' and ``abstract entity'' with labels starting with LOC are for genes.

The program also outputs an error file, but no errors were encountered.

\medskip

The Python program AllNumbers.py in the ancillary files
first queries Wikidata to find the disjoint pairs in the disjoint union of statements
and then runs a query for each pair that finds the number of culprits for the pair
and the number of disjointness violations at both the subclass and instance level.

The main results are in AllNumbers.csv, which contains one line of output for each disjointness pair that has violations, shown as for AllCulprits.py,
with the number of culprits for the pair,
the total number of subclasses of all the culprits for the pair,
the total number of instances of all the culprits for the pair,
and the total of both subclasses and instances of all the culprits for the pair.

The program also outputs all disjoint pairs, even those with no violations, in AllDisjoints.csv.

\end{document}